\documentclass[sigconf, screen]{acmart}

\AtBeginDocument{%
  \providecommand\BibTeX{{%
    \normalfont B\kern-0.5em{\scshape i\kern-0.25em b}\kern-0.8em\TeX}}}

\copyrightyear{2020}
\acmYear{2020}
\setcopyright{acmlicensed}\acmConference[RecSys '20]{Fourteenth ACM Conference on Recommender Systems}{September 22--26, 2020}{Virtual Event, Brazil}
\acmBooktitle{Fourteenth ACM Conference on Recommender Systems (RecSys '20), September 22--26, 2020, Virtual Event, Brazil}
\acmPrice{15.00}
\acmDOI{10.1145/3383313.3412243}
\acmISBN{978-1-4503-7583-2/20/09}

\usepackage{amsmath,amsfonts,bm}

\def\eqref#1{equation~\ref{#1}}

\def\1{\bm{1}}

\def\vr{{\bm{r}}}

\def\vx{{\bm{x}}}

\def\vz{{\bm{z}}}

\def\vmu{{\bm{\mu}}}
\def\vnu{{\bm{\nu}}}

\def\mF{{\bm{F}}}

\def\mI{{\bm{I}}}

\def\mP{{\bm{P}}}
\def\mQ{{\bm{Q}}}
\def\mR{{\bm{R}}}

\def\mX{{\bm{X}}}

\def\fmX{{\widehat{\bm{X}}}}
\def\fmR{{\widehat{\bm{R}}}}
\def\fvx{{\widehat{\bm{x}}}}
\def\fvr{{\widehat{\bm{r}}}}
\def\matrr{\begin{bmatrix}
\vr_i \\
\fvr_i \\
\end{bmatrix}}
\def\matPF{\begin{bmatrix}
\mP \\
\mF \\
\end{bmatrix}}
\def\matxx{\begin{bmatrix}
\vx_i \\
\fvx_i \\
\end{bmatrix}}

\DeclareMathAlphabet{\mathsfit}{\encodingdefault}{\sfdefault}{m}{sl}
\SetMathAlphabet{\mathsfit}{bold}{\encodingdefault}{\sfdefault}{bx}{n}

\def\sR{{\mathbb{R}}}

\def\emF{{F}}

\def\emP{{P}}
\def\emQ{{Q}}

\newcommand{\ltrain}{\mathcal{L}_{\rm{train}}}
\newcommand{\ladv}{\mathcal{L}_{\rm{adv}}}

\newcommand{\norm}[1]{\left\lVert#1\right\rVert}

\DeclareMathOperator*{\argmin}{arg\,min}

\usepackage[T1]{fontenc}    %
\usepackage{booktabs}       %
\usepackage{amsfonts}       %
\usepackage{nicefrac}       %
\usepackage{microtype}      %

\usepackage{arydshln}
\usepackage{subcaption,verbatim,engord,enumitem,multirow,bm}
\usepackage{algorithm,algorithmic}
\usepackage{amsmath,dsfont}
\usepackage{amssymb}
\usepackage{array}
\usepackage{amscd}
\usepackage{mathtools}
\usepackage{wrapfig}
\usepackage[capitalise]{cleveref}

\begin{document}

\title{Revisiting Adversarially Learned Injection Attacks Against Recommender Systems}

\author{Jiaxi Tang}
\authornote{Now at Google Inc., work done when he was a student at Simon Fraser University.}
\affiliation{%
  \institution{Simon Fraser University}
  \city{British Columbia} 
  \state{Canada} 
}
\email{jiaxit@sfu.ca}

\author{Hongyi Wen}
\affiliation{%
  \institution{Cornell Tech, Cornell University}
  \city{New York} 
  \state{NY, USA} 
}
\email{hw557@cornell.edu}

\author{Ke Wang}
\affiliation{%
  \institution{Simon Fraser University}
  \city{British Columbia} 
  \state{Canada} 
}
\email{wangk@cs.sfu.ca}

\renewcommand{\shortauthors}{Tang, et al.}

\begin{abstract}
Recommender systems play an important role in modern information and e-commerce applications. While increasing research is dedicated to improving the relevance and diversity of the recommendations, the potential risks of state-of-the-art recommendation models are under-explored, that is, these models could be subject to attacks from malicious third parties, through injecting fake user interactions to achieve their purposes. This paper revisits the adversarially-learned injection attack problem, where the injected fake user `behaviors' are learned locally by the attackers with their own model -- one that is potentially different from the model under attack, but shares similar properties to allow attack transfer. We found that most existing works in literature suffer from two major limitations: (1) they do not solve the optimization problem precisely, making the attack less harmful than it could be, (2) they assume perfect knowledge for the attack, causing the lack of understanding for realistic attack capabilities. We demonstrate that the exact solution for generating fake users as an optimization problem could lead to a much larger impact. Our experiments on a real-world dataset reveal important properties of the attack, including attack transferability and its limitations. These findings can inspire useful defensive methods against this possible existing attack.
\end{abstract}



\maketitle

\section{Introduction}\label{sec:intro}
A good recommender system is a key factor to users' information seeking experience as it enables better content discovery and more accurate information retrieval. Over the past decade, most work aims to improve the utility/accuracy of recommendation models. Many methods have been developed, such as neighborhood-based methods~\cite{sarwar2001item}, factorization-based approaches~\cite{koren2009matrix,rendle2009bpr}, and the more recent deep neural network~(\emph{a.k.a} deep learning) models~\cite{cheng2016wide,he2017neural}. However, due to the reliance on user contributed judgments and subjective rating data~\cite{burke2015robust}, recommender systems can be misused and attacked with malicious purposes. Once this happens, the credibility of a recommender system will be largely affected, which could lead to a significant economic loss.

\subsection{Injection Attack against Recommender Systems}

\begin{figure*}[t!]  %
\centering
\includegraphics[scale=0.45]{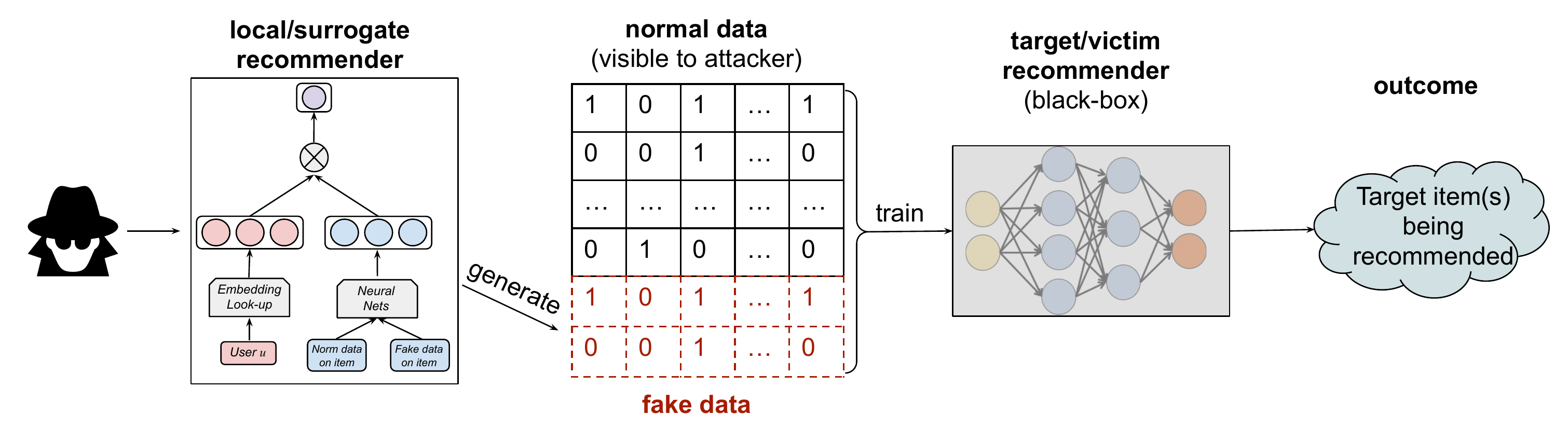}
\caption{An illustration of the threat model for injection attack against recommendation models. This assumes the dataset is available to the attacker but the target model is unknown. To achieve their malicious goals, e.g., influencing certain item(s)' availability of being recommended, the attacker will craft fake user profiles locally with a surrogate model and inject them to the target recommender before it is trained.}
\label{fig:diagram}
\end{figure*}

In this work, we focus on \textit{injection attack} (a.k.a. data poisoning attack) as illustrated in~\cref{fig:diagram}, where the malicious party has knowledge about the data used by a recommender system (\emph{e.g.}, by crawling the publicly available data) and creates fake user profiles with carefully chosen item preferences (\emph{e.g.,} clicks) to influence the recommender with malicious goals. Assuming such knowledge about the data is reasonable, for example: users' ratings and reviews on Amazon's product are public\footnote{\url{http://jmcauley.ucsd.edu/data/amazon}}, which account for the personalized product recommendations;
users' social relationship (followings and followers) on Twitter is public, which influence friend recommendations; users' answers on questions and upvotes on answers are public on Quora, which constitute the personalized feed.
As long as there are enough incentives, the availability of a dataset from certain platforms is \emph{not unreachable} to the malicious party.
It was reported that every one-star increase in user ratings on certain product can lead to a 5 to 9 percent increase in product seller's revenue\footnote{\url{https://www.forbes.com/sites/ryanerskine/2018/05/15/you-just-got-attacked-by-fake-1-star-reviews-now-what}}. Therefore, malicious injection attacks are easily motivated and can have huge consequences for a company's bottom line. 

In the literature, early studies on performing injection attack against recommender systems are inspired by heuristics.
To boost certain item(s)' availability of being recommended, \citet{lam2004shilling} proposed to give high ratings for the targeted item and average ratings on other random items; \citet{burke2005limited} further considered to use popular items instead of random items to ensure that fake users can have more neighborhoods thus have more impacts.
However, since these attacks were created heuristically, the threat of injection attacks may not be fully realized.
First, the fake users generated are highly correlated with each other and sometimes even self-forming clusters~\cite{burke2015robust}, making them easily detectable by standard techniques~\cite{mehta2008survey}.
What's more, heuristic methods heavily rely on background knowledge, hence, a method designed for one malicious purpose is hard to be used for another.
Finally, heuristic methods do not directly optimize the adversarial goals, which limits their usability and threat.

Recently, we have witnessed a huge impact of adversarial attacks through adversarial machine learning that optimizes an adversarial objective, 
irrespective of the model type and tasks~\citep{mei2015using}: In web search, an adversary can change web contents to get high search engine's rankings~\cite{castillo2011adversarial}; In crowd-sourcing, an adversary can provide useless answers for profits~\cite{miao2018attack}; In social networks, an adversary can modify node relationships for a desired node property~\cite{zugner2018adversarial}; In image recognition, an adversary can make perturbations on image pixels and have a wanted recognition result~\cite{goodfellow2014adversarial, kurakin2016adversarial}. 

Despite the success of adversarial learning in other domains, there's very sparse research on adopting adversarial learning to attack recommender systems. 
In the security arms race, a limited knowledge of the attack leads to a more dangerous state of existing systems. 
In this work, we aim to revisit this direction by investigating the challenges and limitations of using adversarial learning to attack recommender systems. In the next two sections, we will have the same viewpoint as adversaries to understand how the attack can be performed. 
After defining the threat model in~\cref{sec:background}, we found that existing works do not solve the problem properly, causing the attack less powerful than it could have been.
In~\cref{sec:solution}, we propose a more precise but less efficient solution to generate the attack, accompanied by two efficient approximations to the solution. In~\cref{sec:expr}, we explore the attack's impact on a real-world dataset and identify its weakness and find the clues of such attacks. We hope that these findings can help us better understand the attack thus develop defensive techniques.

\section{Background}\label{sec:background}
In this section, we first cover essential basics of the recommendation task and define the threat model. Some notations will also be introduced to facilitate the presentation. Then we briefly revisit existing solutions and their limitations, which encourage us to propose a more precise approach in the next section.

\subsection{Recommendation Goal}
The goal of system under-attack is to recommend relevant items to cater users' needs. In such a system, there is a set of users $\mathcal{U} = \{u_1, u_2,..,u_{|\mathcal{U}|}\}$, a set of items $\mathcal{I} = \{i_1, i_2,..,i_{|\mathcal{I}|}\}$ (e.g., products, videos, venues, etc.), and feedback/interaction data (e.g., user purchased a product, watched a video, checked-in at a venue).
The feedback that users left in the system can have different types.
It can be either \emph{explicit} (i.e., the user explicitly shows how she/he likes an item, such as giving a five star rating) or \emph{implicit} (i.e., a signal implicitly reflect user's positive preferences, such as purchasing a product), but the later is much more prevalent than the explicit feedback in real systems~\cite{pan2008one, hu2008collaborative} thus is considered in this work.
We use $\mX \in \{0,1\}^{|\mathcal{U}| \times |\mathcal{I}|}$ to denote the binarized implicit data, with 1 for a positive feedback and 0 for an unknown entry.
A recommendation model built on users' historical data can make predictions $\mR \in \sR^{|\mathcal{U}| \times |\mathcal{I}|}$.
The learning objective of a recommendation model is to provide relevant items of each user with the highest predicted relevance scores.

\subsection{Threat Model}\label{sec:background_threat_model}
The threat model is illustrated in Figure~\ref{fig:diagram}. To attack the target recommender deployed in the system (a.k.a. \textit{victim model}), the attacker will use their own local model (a.k.a. \textit{surrogate model}), to craft fake users and inject them into the original training data of victim model. Below, we elaborate the threat model from several perspectives.

\noindent{\textbf{Attacker's goal.}}
The adversary's goal can be either \emph{non-targeted}, aiming to hamper the effectiveness of the recommendation model by forcing it to make bad recommendations, or \emph{targeted}, where the adversary wishes to increase or decrease a target item(s)' availability of being recommended.
Similar to most other works~\cite{lam2004shilling, burke2005limited,fang2018poisoning,fang2020influence} in literature, we mainly focus on \emph{targeted attack}, which is the most common case under recommendation context: attackers want to influence normal users' recommendations for profits.
Specifically to targeted attack, we consider the \emph{promote (or push) attack}: given a target item, attacker's goal is to increase its chance of being recommended by victim model. Alternatively, there is \emph{nuke attack}, where attackers aim to ``nuke'' a target item, make it less able to get recommended. Although we don't explicitly discuss the nuke attack in this work, similar techniques can be used.

\noindent{\textbf{Attacker's knowledge.}}
We assume the attacker has (full or partial) knowledge about the dataset used to train the target (victim) recommendation model. Because user feedback is public in many systems (e.g., followings and followers on Twitter), this is a reasonable assumption for a worst-case attack. Any knowledge about victim model (e.g., model parameters, model type, etc) is optional, because attackers can first attack their own local (surrogate) model with the poison fake users, hopeing that these fake users can be also used to attack the target (victim) model. This kind of \emph{attack transfer} is possible if two models share similar properties~\citep{papernot2016transferability}.

\noindent{\textbf{Attacker's capability.}}
As shown in~\cref{fig:diagram}, the attacker will learn fake users through a surrogate model and inject their rating profiles to the training set of the victim model. The attacker will achieve their malicious goal after the victim model consumes these fake users. This suggests that the attack happens at training time of the victim model, instead of test time. The later is known as evasion attack and is more commonly studied in adversarial learning~\cite{goodfellow2014adversarial, kurakin2016adversarial,papernot2016transferability,papernot2017practical}. While in recommendation, test time attack requires attackers to hack into other normal users' accounts and change their preferences, which is a cybersecurity issue and is beyond the scope of our work. On the other hand, allowing adversaries to create fake users and let them be consumed (trained) by the victim model is a more practical attack scenario.

\subsection{Adversarial Injection Attack: a bi-level optimization problem}
Different from heuristic approaches~\cite{lam2004shilling, burke2005limited}, the injection attack considered in this paper directly \emph{learns} fake user behaviors as an optimization problem.
Recall that the adversaries will learn fake users to attack their own surrogate model as a first step.
Given a well-trained surrogate model that is under-attack and a set of fake users $\mathcal{V} = \{v_1, v_2,..,v_{|\mathcal{V}|}\}$, the fake data $\fmX \in \{0,1\}^{|\mathcal{V}| \times |\mathcal{I}|}$ will be learned to optimize an adversarial objective function $\ladv$
\begin{equation}\label{eqn:optim_outer}
\min_{\fmX}~\ladv(\mR_{\theta^*}),
\end{equation}
\begin{equation}\label{eqn:optim_inner}
\textrm{subject to}\quad \theta^* = \argmin_\theta \big ( \ltrain(\mX, \mR_{\theta}) + \ltrain(\fmX, \fmR_{\theta}) \big ),
\end{equation}
where $\theta$ denotes a set of surrogate model's parameters, $\mR_\theta$ is surrogate model's predictions on normal users with parameter $\theta$ and $\ltrain$ denotes surrogate model's training objective.
As shown, one optimization problem (i.e., \cref{eqn:optim_inner}, called \emph{inner objective}) is embedded (nested) within another (i.e., \cref{eqn:optim_outer},  called \emph{outer objective}), this forms a bi-level optimization problem~\cite{colson2007overview}. In machine learning, the bi-level optimization has been considered for hyperparamter optimization~\cite{franceschi2018bilevel,chen2019lambdaopt}, few-shot learning and meta-learning~\cite{finn2017model,rajeswaran2019meta, grefenstette2019generalized}. We formulate the learned injection attack as a bi-level optimization problem because of the definition of the threat model introduced in~\cref{sec:background_threat_model}, although we found this formulation is not explicitly discussed in literature.
The inner objective\footnote{Note that although here we separate the training objective for normal data and for fake data to have a clear presentation, in reality, these two data are arbitrarily mixed together and cannot be distinguished.} shows that after fake data $\fmX$ are injected, the surrogate model will first consume them (i.e., train from scratch with the poisoned dataset), we then obtain the trained model parameters $\theta^*$.
The outer objective shows that after fake data are consumed, we can achieve the malicious goal defined on normal user's predictions $\mR_{\theta^*}$.
Note that different adversarial objectives can be used for different malicious purposes. In this paper, we focus on promoting target item $k$ to all normal users, so an exemplary adversarial objective can be the cross-entropy loss:
\begin{equation}\label{eqn:obj_ce}
\ladv(\mR) = - \sum_{u \in \mathcal{U}} \log \left( \frac{\exp(r_{uk})}{\sum_{i \in \mathcal{I}} \exp(r_{ui})} \right) .
\end{equation}
The objective will be minimized if normal user's prediction on target item $k$ is greater than other items, so that the malicious goal of promoting target item will be achieved, as a result.

\begin{algorithm}[t]
\caption{Learning fake user data with Gradient Descent}
\begin{algorithmic}[1]\label{alg:optim_gd}
\STATE \textbf{Input:} Normal user data $\mX$; learning rate for inner and outer objective: $\alpha$ and $\eta$; max iteration for inner and outer objective: $L$ and $T$.
\STATE \textbf{Output:} Learned fake user data for malicious goal.
\STATE Initialize fake data $\fmX^{(0)}$ and surrogate model parameters $\theta^{(0)}$
\FOR{$t=1$ to $T$}
\FOR{$l=1$ to $L$}
\STATE Optimize inner objective with SGD: $\theta^{(l)} \gets \theta^{(l-1)} - \alpha \cdot \nabla_{\theta} \big ( \ltrain(\mX, \mR_{\theta^{(l-1)}}) + \ltrain(\fmX^{(t)}, \fmR^{(t)}_{\theta^{(l-1)}}) \big )$
\ENDFOR
\STATE Evaluate $\mathcal{L}_\textrm{adv}(\mR_{\theta^{(L)}})$ and compute gradients $\nabla_{\fmX}\ladv$
\STATE Update fake data:
$\fmX^{(t)} = \textrm{Proj}_{\Lambda} \big ( \fmX^{(t-1)} - \eta \cdot \nabla_{\fmX}\ladv \big )$
\ENDFOR
\STATE \textbf{Return:} $\fmX^{(T)}$
\end{algorithmic}
\end{algorithm}
To solve the bi-level optimization problem in Eqs.~(\ref{eqn:optim_outer}) to (\ref{eqn:optim_inner}), one could try every possible $\fmX$, obtain the associated $\theta^*$ and evaluate $\ladv(\mR_{\theta^*})$. But the search space is exponentially large as $2^{|\mathcal{V}| \times |\mathcal{I}|}$. So this brute-force approach can hardly be used with limited resources.
A more computationally-efficient way is to use gradient-based approaches, such as Gradient Descent, to iteratively update fake data $\fmX$ with gradient $\nabla_{\fmX}\ladv$, which we formally present in Algorithm~\ref{alg:optim_gd}. 
At each iteration $t \in \{1,...,T\}$ for updating fake data, we first retrain the surrogate model by performing parameter updates for $L$ iterations (line 7). Then we update fake data (line 10) with Projected Gradient Descent (PGD), with $\rm{Proj}_{\Lambda}(\cdot)$ as the projection operator that projects the fake data onto feasible set (i.e., $\hat{x}_{vi} \in \{0,1\}$ in our case). After the final iteration $T$ where fake data $\fmX^{(T)}$ is learned to minimize $\ladv$, they are able to attack the surrogate model, and we hope they can also attack the target (victim) model in a similar way: once trained on the poisoned dataset, it will cause a small adversarial loss $\ladv$.

\subsection{Limitations in Existing Studies and Our Contributions}
There are a few studies~\cite{christakopoulou2019adversarial, li2016data, fang2018poisoning, fang2020influence} in the literature tried to regard the injection attack as an optimization problem and learn fake data for adversarial goals. However, we found there exist two major limitations in existing works. In below, we illustrate these limitations, as well as our contributions in this work.

\textit{Lacking exactness in gradient computation.}
As we can see from \cref{alg:optim_gd}, solving the inner objective for surrogate model training is simple and conventional, while the challenge comes from obtaining the adversarial gradient $\nabla_{\fmX}\ladv$ to update fake data.
In the literature, existing works either tried to estimate this gradient~\cite{christakopoulou2019adversarial}, or tried to directly compute it~\cite{li2016data, fang2018poisoning, fang2020influence}. But under the problem formulation in Eqs.~(\ref{eqn:optim_outer}) to (\ref{eqn:optim_inner}), they all lack exactness in gradient computation.
More specifically, by applying chain rule, the exact adversarial gradient can be written as:
\begin{equation}\label{eqn:adv_grad}
\nabla_{\fmX}\ladv = \frac{\partial\ladv}{\partial \fmX} + 
\frac{\partial\ladv}{\partial \theta^*} \cdot \frac{\partial\theta^*}{\partial \fmX}.
\end{equation}
The first part (partial derivative $\partial\ladv / \partial \fmX$) assumes $\fmX$ is independent to other variables, while the second part suggests $\theta^*$ can be also a function containing $\fmX$.
Among all existing studies~\cite{christakopoulou2019adversarial, li2016data, fang2018poisoning, fang2020influence}, we found the second part in~\cref{eqn:adv_grad} has been completely ignored. This suggests the final surrogate model parameters $\theta^*$ is independent from fake data $\fmX$, but it is obviously incorrect. 
In~\cref{sec:solution}, we show the first part doesn't exist in many surrogate models and when it exists, the second part also contributes significantly to the total gradient.
This suggests the approximation of gradient $\nabla_{\fmX}\ladv$ with its partial derivative is largely biased, therefore can lead to sub-optimal solutions.

\textit{Our contributions.}
In Section~\ref{sec:solution}, we present the computation of exact adversarial gradient in~\cref{eqn:adv_grad} and show two efficient ways to approximate this gradient. On a synthetic and a real-world dataset, we empirically demonstrate the effectiveness of both approaches and the undesirable results of only computing the partial derivative for approximation, as used in existing works~\cite{christakopoulou2019adversarial, li2016data, fang2018poisoning, fang2020influence}.

\paragraph{Lacking vital experimental studies.}
Another major limitation in all previous experimental studies~\cite{christakopoulou2019adversarial, li2016data, fang2018poisoning, fang2020influence}, is that the target model is set to be identical to the surrogate model, which is known as ``white-box attack''. One could think this is an extreme case where target model is visible to the adversary and this can serve as a upper bound of the attack capability. Our analysis shows that this attack may not be carried over to the realistic case where the target model is different from the surrogate model. Knowing this, the attacker would design the more effective attack by learning fake user data using a surrogate model that is tranferable to a different target model. The ultimate goal is to defend against attacks, which requires considering more practical settings and understanding of the attack's characteristics and limitations, in order to inspire better defensive strategies.

\textit{Our contributions.}
In Section~\ref{sec:expr}, we leverage a user-venue check-ins dataset to study how attack crafted from one surrogate model can transfer to another victim models and examine the key factors that influence the transferability.
More importantly, we analyze the limitations of this adversarially-learned injection attack, which could inspire useful defensive techniques.

\section{Solving the bi-level optimization problem}\label{sec:solution}
In this section, we focus on line 9 of~\cref{alg:optim_gd} and describe how to compute the adversarial gradient in \cref{eqn:adv_grad} exactly  (in~\cref{sec:solution_exact}) and provide two approximated solutions (in~\cref{sec:solution_approx1} and~\cref{sec:solution_approx2}). First of all, we will use Weighted Regularized Matrix Factorization (WRMF)~\cite{hu2008collaborative, pan2008one}, a fundamental and representative factorization-based model for recommendations with implicit feedback, as an example of the surrogate model and demonstrate how the exact adversarial gradient can be computed.
However, the exact gradient computation is neither time-efficient nor resource-efficient. We thus introduce two orthogonal ways to approximate the gradient computation to achieve a good balance between effectiveness and efficiency. On a synthetic toy dataset, we empirically evaluate how good are the approximated solutions.

Before diving into details, we briefly introduce the WRMF model and the toy dataset used in this section. In WRMF, a set of user latent factors $\mP \in \sR^{|\mathcal{U}| \times K}$ and item latent factors $\mQ \in \sR^{|\mathcal{I}| \times K}$ are used to make predictions $\mR=\mP \mQ^\top$ on normal data $\mX$.
When fake data $\fmX$ are injected, let $\mF \in \sR^{|\mathcal{V}| \times K}$ denotes the user latent factors and $\fmR=\mF \mQ^\top$ denotes the predictions for fake users.
Under this formulation, $\theta=\{\mP,\mQ,\mF\}$ and the surrogate training objective is:
\begin{equation}\label{eq:wrmf_obj}
\begin{aligned}
&\ltrain(\mX, \mR_{\theta}) + \ltrain(\fmX, \fmR_{\theta})=\\
&\qquad \sum_{u,i} w_{ui}(x_{ui} - \emP_u^\top \emQ_i)^2 + \sum_{v,i} w_{vi} (\hat{x}_{vi} - \emF_v^\top \emQ_i)^2\\
&\qquad + \lambda \left (\norm{\mP}^2+ \norm{\mF}^2 + \norm{\mQ}^2 \right ),
\end{aligned}
\end{equation}
where $w_{ui}$ and $w_{vi}$ are instance weights to differentiate observed and missing feedback from the normal and fake data, respectively (e.g., $w_{ui}=2$ when $x_{ui}=1$ and $w_{ui}=1$ when $x_{ui}=0$, similar for $w_{vi}$), $\lambda$ is the hyperparameter to control model complexity.

\textbf{Synthetic data.}
To facilitate our understanding of exact and approximated solutions for computing adversarial gradient, we synthesize a toy datset that is more controllable. Specifically, each data point in $\mX$ is generated by $x=\vmu \vnu^\top$, where both $\vnu \in \sR^d$ and $\vmu\in \sR^d$ are sampled from $\mathcal{N}_d(\mathbf{0}, \mathbf{I})$ with $d<<\min(|\mathcal{U}|, |\mathcal{V}|)$. By generating data point for $\forall x \in \mX$, the synthesized dataset is presumably to have low-rank, similar to other real-world recommendation datsets. Lastly, we binarize $\mX$ to transform it into implicit feedback data by setting a threshold $\epsilon$. By controlling the value of $(|\mathcal{U}|,|\mathcal{V}|,d,\epsilon)$, we are able to have arbitrary-size synthetic datasets with different ranks and sparsity levels.

\subsection{Exact Solution}\label{sec:solution_exact}
In this subsection, we compute the exact adversarial gradient $\nabla_{\fmX}\ladv$ when WRMF is used as the surrogate model.
It's worth noting that, WRMF's predictions on normal users $\mR=\mP \mQ^\top$ are not directly dependent on fake data $\fmX$, therefore $\ladv(\mR)$ is not differentiable w.r.t. $\fmX$ or equivalently, we can specify $\frac{\partial\ladv}{\partial \fmX}=0$.
In fact, this is a common case for most embedding-based recommendation models~\cite{he2017neural, hsieh2017collaborative, rendle2009bpr}, in which data ($\mX$ and $\fmX$) are only used for computing training objective (as labels), and are not involved in computing predictions ($\mR$ and $\fmR$).

\begin{figure*}[t!]  %
    \centering
    \begin{subfigure}[b]{0.65\textwidth}
        \includegraphics[width=\textwidth]{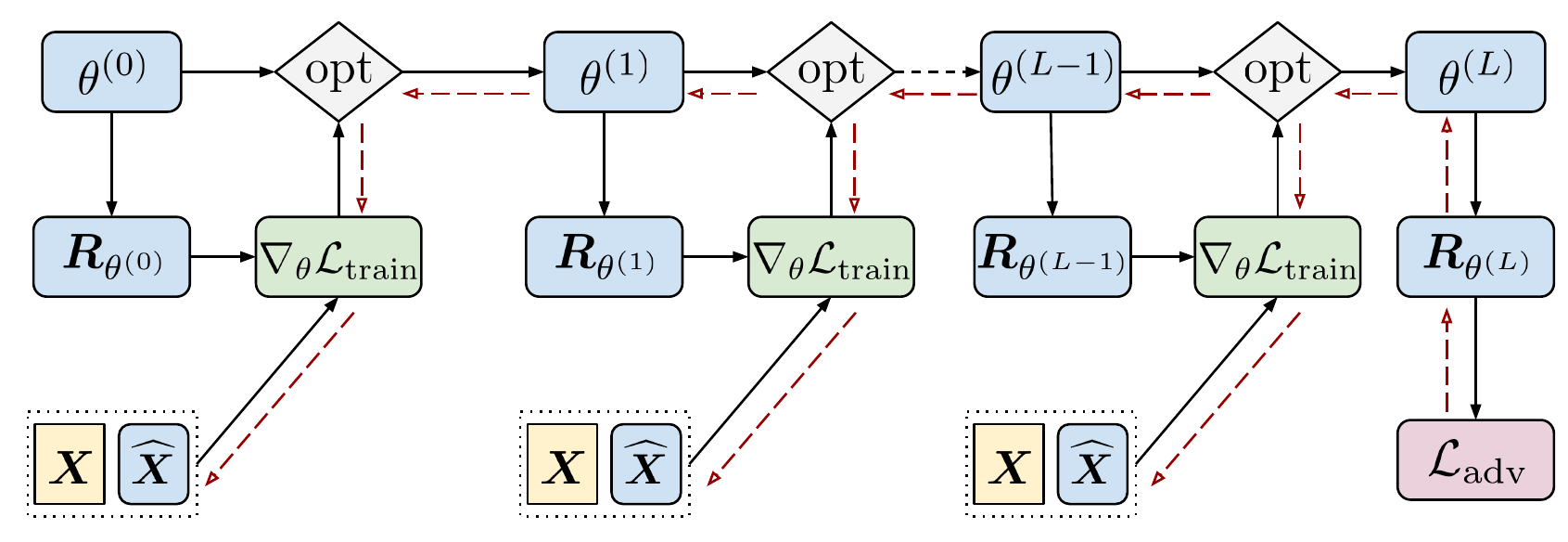}
        \vspace{-2em}
        \caption{}
        \label{fig:comp_graph}
    \end{subfigure}
    \hspace{2em}
    \begin{subfigure}[b]{0.25\textwidth}
        \includegraphics[width=\textwidth]{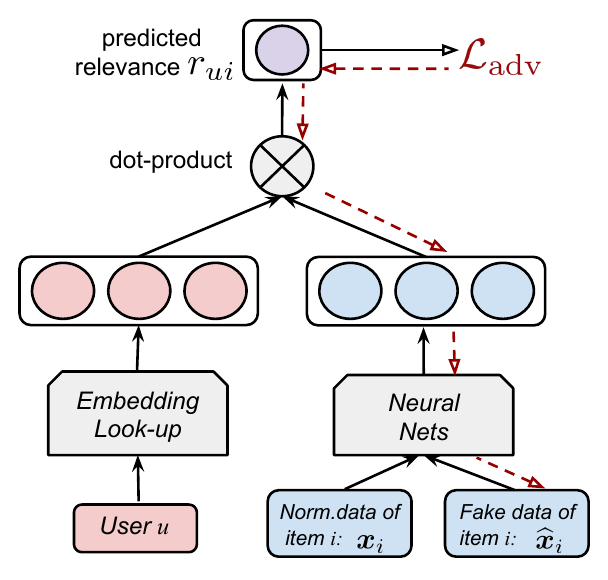}
        \vspace{-2em}
        \caption{}
        \label{fig:attin}
    \end{subfigure}
    \vspace{-1em}
    \caption{(a) Computational graph for a single calculation of adversarial objective $\ladv$ with surrogate model. We use solid black arrow to denote the forward computation flow and use dashed red arrow to denote the gradient backpropagation flow, from $\ladv$ to $\fmX$. (b) The proposed surrogate model in this paper, the model can be also viewed as item-based autoencoder.}
\end{figure*}

Without loss of generality, we assume the inner objective is optimized for only once (i.e., $L=1$), then $\theta^{(1)}=\{\mP^{(1)},\mQ^{(1)},\mF^{(1)}\}$ is the final parameter set used in adversarial objective $\ladv(\mR_{\theta^{(1)}})= \ladv(\mP^{(1)} \cdot \mQ^{(1)})$.
Also, we can formulate $\theta^{(1)}=\textbf{opt}(\theta^{(0)}, \nabla_{\theta} \ltrain)$, here $\textbf{opt}(,)$ denotes the transition function from $\theta^{(l-1)}$ to $\theta^{(l)}$. As in~\cref{alg:optim_gd}, under the context of Stochastic Gradient Descent (SGD), this will become
\begin{equation*}
\begin{aligned}
\theta^{(1)} &= \textbf{opt} \Big( \theta^{(0)}, \nabla_{\theta} \big( \ltrain(\mX, \mR_{\theta^{(0)}}) + \ltrain(\fmX, \fmR_{\theta^{(0)}}) \big) \Big)\\
&= \theta^{(0)} - \alpha \cdot \nabla_{\theta} \big( \ltrain(\mX, \mR_{\theta^{(0)}}) + \ltrain(\fmX, \fmR_{\theta^{(0)}}) \big).
\end{aligned}
\end{equation*}
Now we can easily compute the adversarial gradient $\nabla_{\fmX}\ladv$ when using WRMF and $T=1$, by applying chain rule:
\begin{equation}\label{eqn:dL_dx}
\begin{aligned}
&\nabla_{\fmX}\ladv = \frac{\partial\ladv}{\partial \fmX} + 
\frac{\partial\ladv}{\partial \theta^*} \cdot \frac{\partial\theta^*}{\partial \fmX}
=0 + \frac{\partial\ladv}{\partial \theta^{(1)}} \cdot \frac{\partial\theta^{(1)}}{\partial \fmX}\\
& = \frac{\partial\ladv}{\partial \theta^{(1)}} \cdot \Big( -\alpha \nabla_{\fmX} \nabla_{\theta} \big( \ltrain(\mX, \mR_{\theta^{(0)}}) + \ltrain(\fmX, \fmR_{\theta^{(0)}}) \big) \Big).
\end{aligned}
\end{equation}
Similarly, when $T>1$, we just need to accumulate the gradient 
\begin{equation}\label{eqn:grad_sum_exact}
\nabla_{\fmX}\ladv=\sum_{l \in [1,L]} \frac{\partial\ladv}{\partial \theta^{(l)}} \cdot \frac{\partial\theta^{(l)}}{\partial \fmX}.
\end{equation}
In the above summation, $\frac{\partial\theta^{(l)}}{\partial \fmX}$ is trivial to obtain, as shown in~\cref{eqn:dL_dx}, while $\frac{\partial\ladv}{\partial \theta^{(l)}}$ can be done in a sequential manner. That is, after having $\frac{\partial\ladv}{\partial \theta^{(l+1)}}$ we can acquire $\frac{\partial\ladv}{\partial \theta^{(l)}}$ by
\begin{equation}\label{eqn:dtheta_dtheta}
\begin{aligned}
&\frac{\partial\ladv}{\partial \theta^{(l)}} = \frac{\partial\ladv}{\partial \theta^{(l+1)}} \cdot \frac{\partial \theta^{(l+1)}}{\partial \theta^{(l)}},\quad\text{where}\\ &\frac{\partial \theta^{(l+1)}}{\partial \theta^{(l)}}=\Big( 1 -\alpha \nabla_{\theta} \nabla_{\theta} \big( \ltrain(\mX, \mR_{\theta^{(l)}}) + \ltrain(\fmX, \fmR_{\theta^{(l)}}) \big) \Big).
\end{aligned}
\end{equation}
In Figure~\ref{fig:comp_graph}, we show the overall computational graph of the procedure for calculating the adversarial gradient $\nabla_{\fmX}\ladv$. Note that this procedure applies to most embedding-based recommendation models, not limit to WRMF, if their inner objectives are also optimized with SGD (or variants of SGD, such as Adam~\cite{kingma2014adam}).
It's worth mentioning that despite the differentiations showed in~\cref{eqn:dL_dx} and~\cref{eqn:dtheta_dtheta} contain the computation of Hessian matrix, \emph{automatic differentiation}~\cite{grefenstette2019generalized} provides us a convenient way to solve them.

To study the effectiveness of attack using the above exact solution, we design a proof-of-concept experiment on a synthetic data set to learn fake user data using WRMF as the surrogate model under a white-box attack setting.

\begin{figure*}[t!]  %
\centering
\includegraphics[width=0.7\textwidth]{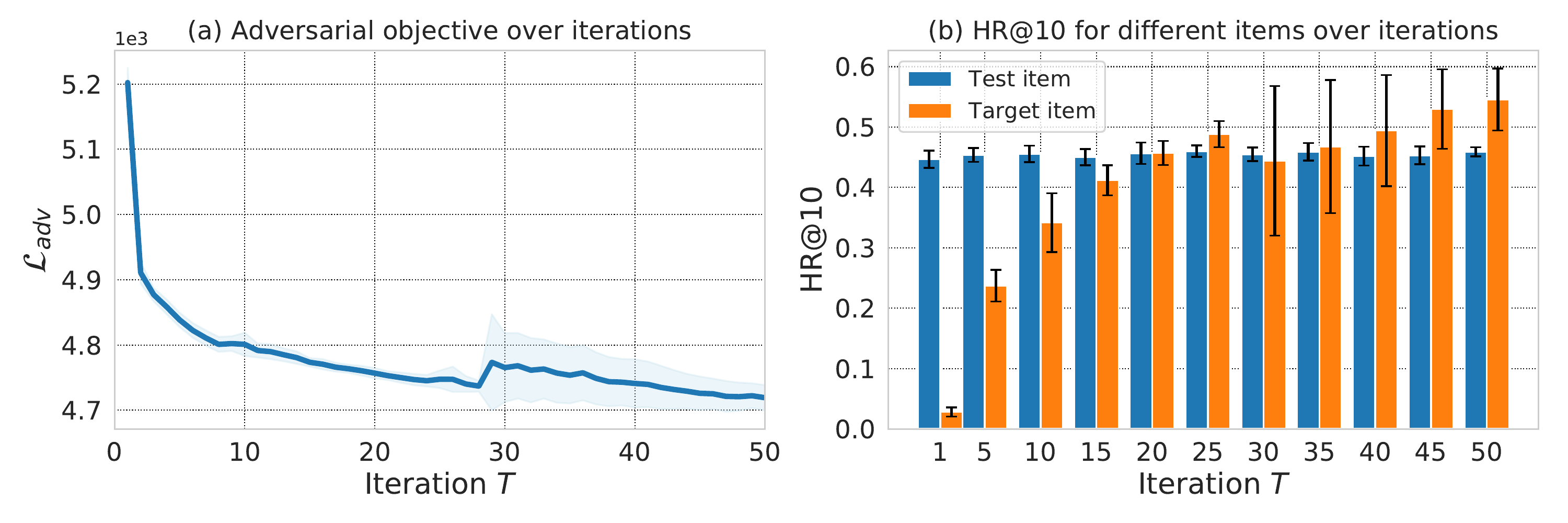}
\vspace{-1.5em}
\caption{On our synthesized toy dataset, we report the mean and standard deviation for (a) adversarial objective and (b) HR@10 for test and target item over iteration $T$. The experiment is repeated for 5 times with different initializations.}
\label{fig:syn_exact}
\end{figure*}

\textbf{Setup.}
Based on the synthetic data generation method we described previously, we create a toy dataset with 900 normal users, 100 fake users and 300 items. That is, $|\mathcal{U}|=900, |\mathcal{V}|=100$, and $|\mathcal{I}|=300$ in the following experiments. When generating data, we set its `rank' $d$ to be 20 and binarization threshold $\epsilon$ to be 5, the data sparsity is 88\% under this setting. The model we use is WRMF with latent dimensionality $K=16$ and with its training objective showed in \cref{eq:wrmf_obj}. In terms of instance weight, we empirically set $w_{ui}=20$ when $x_{ui}=1$ and $w_{ui}=1$ when $x_{ui}=0$, as this gives the best recommendation performance. Same weight also applies for $w_{vi}$. Finally, surrogate model (i.e., the inner objective) is optimized with Adam for 100 iterations ($L=100$) and fake data (i.e., the outer objective) is optimized for 50 iterations ($T=50$).

\textbf{Implementation details.}
Recall that after updating the fake data with the exact adversarial gradients, we have to project them onto feasible region (i.e., $\hat{x}_{vi} \in \{0,1\}$). To achieve this, the most straightforward way is using a threshold:
$$
\rm{Proj}_{\Lambda}(x) =
\left\{
	\begin{array}{ll}
		1  & \mbox{if } x \ge \rho, \\
		0 & \mbox{else }.
	\end{array}
\right.
$$
Empirically, one can apply grid-search for the value of $\rho$ for larger attack influence. In our experiments, we use $\rho=0.2$.

\textbf{Results.}
First of all, to evaluate the performance of the surrogate model as well as the performance of learned injection attack, we use \textit{hit ratio} truncated at 10 (HR@10) as the metric.
To evaluate surrogate model's recommendation performance, we randomly reserve 1 interacted item per user (denoted as \emph{test item}), measure HR@10 on this test item and average it for all normal users. To evaluate the attack performance, we use the same strategy but measure HR@10 on \emph{target item}.
\cref{fig:syn_exact} shows the results for both performance we care about. 
From \cref{fig:syn_exact}(a), we can see when using WRMF as the surrogate model and its exact adversarial gradients to solve the bi-level optimization problem in Eqs.~(\ref{eqn:optim_outer}) to (\ref{eqn:optim_inner}), the adversarial objective $\ladv$ is successfully minimized over iterations.
From \cref{fig:syn_exact}(b), we notice the attack performance (i.e., HR@10 on target item) is also getting better along with $\ladv$. More interestingly, we found the attack doesn't change much to WRMF's recommendation performance (i.e., HR@10 on test item).

With the experiment showed above, we illustrate the effectiveness of the learned injection attack when (1) using WRMF with Adam as surrogate model and (2) using the exact adversarial gradients accumulated for all iterations $l \in \{1,..,L\}$ in inner objective optimization. Ideally, we can substitute WRMF with other type of surrogate models (or an ensemble of several models) to achieve the best attack performance against most victim models.
This effective approach has been tried in other domains~\cite{munoz2017towards}, but remains under-explored for injection attack against recommender systems.

Nevertheless, this method has a big drawback due to its high complexity in both time and space. Originally, when optimizing the inner objective, we perform forward and backward propagation to update surrogate model parameters $\theta$, and we only keep the latest values of these parameters.
While if we want to compute the exact gradient (as shown in~\cref{fig:comp_graph}), we need to store the parameter values $\theta^{(l)}$ for each single iteration $l \in \{1,..,L\}$. So \emph{the space complexity grows linearly with the number of total iterations $L$}. That is, if surrogate model size $|\theta|=m$, then a total of $O(Lm)$ space is needed. As for time complexity, we need extra time to compute $\frac{\partial \theta^{(l+1)}}{\partial \theta^{(l)}}$ and $\frac{\partial\theta^{(l)}}{\partial \fmX}$ for each $l \in \{1,..,L\}$. According to the reverse-mode algorithmic differentiation~\cite{baydin2017automatic}, time complexities of both of these computations are proportional to the model size $m$. Thus \emph{the time complexity also grows linearly with $L$}, and an additional $O(Lm)$ time is needed to have all the gradients accumulated, \emph{for a single update of fake data}. As a result, computing the exact gradient $\nabla_{\fmX}\ladv$ is impractical when having a large surrogate model optimized for many iterations, which is very common in real-world recommendation scenarios.
That's why in the next, we show two approximations of the exact gradient $\nabla_{\fmX}\ladv$.

\subsection{Approximated Solution \romannumeral 1: Unrolling fewer steps}\label{sec:solution_approx1}

A straightforward solution is unrolling fewer steps when accumulating $\frac{\partial\ladv}{\partial \theta^{(l)}} \cdot \frac{\partial\theta^{(l)}}{\partial \fmX}, \forall l \in \{L, L-1, ..., 1\}$. When computing the exact gradient with~\cref{eqn:grad_sum_exact}, one can sum the gradient from $l=L$ back to $l=L-\tau$, instead of $l=1$. In other words, ~\cref{eqn:grad_sum_exact} will be approximated as:
\begin{equation}\label{eqn:grad_approx1}
\nabla_{\fmX}\ladv \approx \sum_{l \in [L-\tau,L]} \frac{\partial\ladv}{\partial \theta^{(l)}} \cdot \frac{\partial\theta^{(l)}}{\partial \fmX},
\end{equation}
where $\tau \in [1,L]$ denotes the unroll steps.
This requires we keep surrogate model parameters only for the last $\tau$ steps, and backpropagate adversarial gradients only within last $\tau$ steps. Therefore, it reduces both time and space complexity from $O(L m)$ to $O(\tau m)$.
We can choose the unroll steps $\tau$ according to the available resources, but theoretically, a larger $\tau$ leads to a better approximation.

\begin{figure*}[t!]  %
\centering
\includegraphics[width=0.7\textwidth]{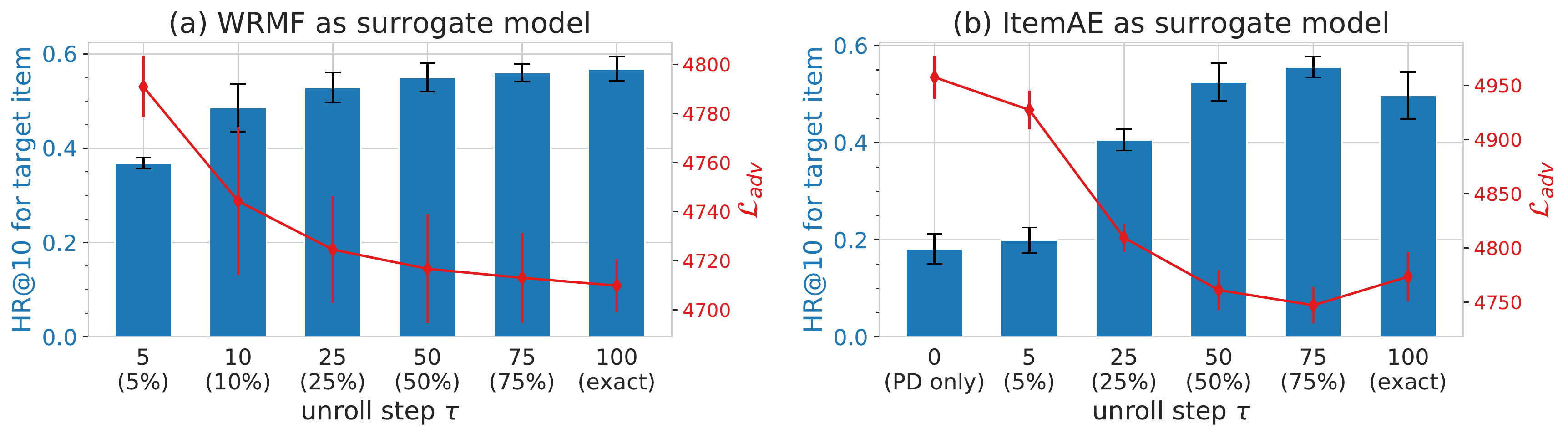}
\vspace{-1.5em}
\caption{On our synthesized toy dataset, we vary the unroll step $\tau$ and report the mean and standard deviation of HR@10 for target item (blue bars) and adversarial objective (red lines). In subfigure (a), WRMF with Adam is used as the surrogate model, while in subfigure (b) ItemAE is used as the surrogate model. Experiments are repeated for 5 times with different initializations.}
\label{fig:syn_approx}
\vspace{-2em}
\end{figure*}

\textbf{Results.}
In~\cref{fig:syn_approx}(a), we show the results of the same experiment in previous subsection, this time varying the number of unroll steps $\tau$. When $\tau=100$, we have the same results as in~\cref{fig:syn_exact}. When using WRMF (optimized with Adam) as surrogate model, we can get better attack performance (measured by a higher HR@10 for target item or a lower $\ladv$) when we unroll more steps, as expected. Notably, even unrolling a few steps (e.g., \%5 of total steps), we can achieve a reasonable attack performance with an approximation factor of 0.65 (0.368 vs. 0.568 in HR@10 for target item). Though promising results are achieved on this synthetic dataset, we'd like to point out the approximation factor is not guaranteed and can vary from different datasets and from different surrogate models.

\subsection{Approximated Solution \romannumeral 2: Using special surrogate models}\label{sec:solution_approx2}
By unrolling fewer steps, we still need extra time and space. While in this subsection, we revisit another way adopted by existing works~\cite{li2016data, fang2020influence}, to approximate the gradient by its partial derivatives: 
\begin{equation}\label{eqn:grad_approx2}
\nabla_{\fmX}\ladv \approx \frac{\partial\ladv}{\partial \fmX}.
\end{equation}
Using this approximation, we don't need to unroll any step (or $\tau=0$) and it requires almost no extra time or space.

Recall that adversarial objective $\ladv$ is defined as a function of predictions on normal data $\mR$ (see \cref{eqn:obj_ce} for an example).
And in~\cref{sec:solution_exact}, we equating $\frac{\partial\ladv}{\partial \fmX}=0$, since when optimized with SGD-based approaches, WRMF's predictions are not depend on the data, neither $\mX$ nor $\fmX$, but only dependent on its parameters $\mR=\mP^\top \mQ$. However,~\citet{li2016data} showed when WRMF is optimized with alternating least square (ALS, or block coordinate descent), then its predictions are explicitly condition on the data, thus we can have non-zero partial derivatives.
In short, when the training of WRMF with ALS is converged, its corresponding predictions for $\mR$ and $\fmR$ are:
$$
\matrr = \matPF  \Big(\matPF^\top\cdot~~\matPF + \lambda \mI \Big)^{-1} \matPF^\top \matxx,
$$
where $\vr_i$ and $\fvr_i$ are the $i$-th column of $\mR$ and $\fmR$, similar denotation also applies to $\vx_i$ and $\fvx_i$. Holding $\mP$ and $\mF$ as constant, we can then compute the desired partial derivatives 
$\frac{\partial\ladv}{\partial \fmX}=\frac{\partial\ladv}{\partial \mR} \cdot \frac{\partial\mR}{\partial \fmX}$.
Using the similar idea,~\citet{fang2018poisoning} showed the partial derivatives exist when using random walk with restart (RWR) as surrogate model. However, a limitation of these approaches is that both WRMF solved with ALS and RWR are not well-supported by existing machine learning libraries (such as TensorFlow~\cite{abadi2016tensorflow} or PyTorch~\cite{paszke2019pytorch}), which are mostly designed for models optimized with SGD-based approaches.
Therefore it's non-trivial to compute all the desired derivatives (both partial derivative in~\cref{eqn:grad_approx2} and the accumulated gradients in~\cref{eqn:grad_approx1}) with \emph{automatic differentiation}~\cite{grefenstette2019generalized}.

In this paper, we offer another surrogate model. Acknowledging the limitation and inspired by previous works, we want the model predictions explicitly condition on data (in other words, $\mX$ and $\fmX$ should be on the computational graph when calculating $\mR$) and is optimized with SGD-based methods. Moreover, this surrogate model is composed with neural networks, thus hopefully, can better transfer attacks to other deep recommenders.
The resulting surrogate model is shown in~\cref{fig:attin}.
Similar to WRMF, we retrieve the user latent representations (red circles) from a user embedding table. While the item latent representations (blue circles) for item $i$ are computed using a feed-forward layer with data $\vx_i$ and $\fvx_i$ as inputs. This makes $\ladv$ differentiable w.r.t. $\fmX$ (red dashed line), thus allows us to have non-zero partial derivatives.

It's worth to note that the model sketched in~\cref{fig:attin} can be also viewed as item-based autoencoder (ItemAE). 
After concatenating normal and fake data for item $i$, i.e., $\vx^+_i = [\vx_i; \fvx_i]$, ItemAE first uses an encoder network $E$ parameterized by $\theta_E$ to project this high-dimensional input to a lower-dimensional latent space $\vz_i=E(\vx^+_i ; \theta_E)$,
where $\vz_i \in \mathbb{R}^K$ is the latent code~(representation) of input $\vx^+_i$ with dimensionality $K$.
ItemAE also uses a decoder network $D$ parameterized by $\theta_D$ to reconstruct the input from its latent code
$\vr^+_i=D(\vz_i ; \theta_D)$. In~\cite{sedhain2015autorec,zhu2019improving}, ItemAE has been investigated for its recommendation performance, but in our paper, we found ItemAE could also help us to obtain a desired partial derivative for learning injection attacks.

\textbf{Results.}
In~\cref{fig:syn_approx}(b), we can see the attack performance when using a ItemAE, optimized by Adam for 100 iterations and with network architecture $(|\mathcal{U}| \rightarrow64 \rightarrow 32 \rightarrow 64 \rightarrow |\mathcal{U}|)$, as surrogate model\footnote{Note that the attack performance in~\cref{fig:syn_approx}(b) is not comparable with the performance in~\cref{fig:syn_approx}(a), as the attacks are evaluated on different victim models.}. 
When using only partial derivative $\frac{\partial\ladv}{\partial \fmX}$ to approximate  $\nabla_{\fmX}\ladv$ (i.e., unroll step $\tau=0$), the attack is weaker, but we benefit from having no extra time and space.
Moreover, since two approximation we discussed are orthogonal to each other, we can still add an unroll step $\tau>0$ for ItemAE to obtain better attack performance, as shown in~\cref{fig:syn_approx}(b). We observe that by incorporating a large unroll step $\tau$, significantly better performance is achieved compared to purely based on the partial derivatives (PD). This demonstrate that the potential of the injection attack is
unfulfilled by ignoring the second term in \cref{eqn:adv_grad}.
Surprisingly, we found using the exact gradient ($\tau=100$) doesn't give the best attack performance. We conjecture this is because the optimization problem in Eqs.~(\ref{eqn:optim_outer}) to (\ref{eqn:optim_inner}) is non-convex, therefore it can be minimized to a better local minima when gentle noises are injected. This is a common phenomena when training non-convex model with SGD, which injects noises but facilitates training.

\subsection{Summary}
Finally, we finish this section by providing a brief summary.
To solve the bi-level optimization problem with gradient-based approaches, the key is to obtain the adversarial gradient $\nabla_{\fmX}\ladv$ in~\cref{eqn:adv_grad}, which is composed with a partial derivative term $\frac{\partial\ladv}{\partial \fmX}$ and accumulated gradients on surrogate model parameters $\theta$. The computation of exact gradient requires high time and space complexity. But we can either use partial derivative to approximate the gradient (special surrogate model is required) or unroll fewer steps when accumulating gradients. Existing works~\cite{christakopoulou2019adversarial, li2016data, fang2018poisoning, fang2020influence} only considered the first approximation method, making the attack weaker than it could be. Underestimating adversary is dangerous in the context of a security arms race~\cite{steinhardt2017certified,biggio2013security}. This is one of the major motivations of this revisiting study.

\begingroup
\renewcommand{\arraystretch}{0.9}
\setlength{\tabcolsep}{11pt}
\begin{table*}[t!]
\caption{The recommendation performance (without attack) and configuration of each model.}
\vspace{-1em}
\begin{tabular}{r|c|c}
\toprule
Model & Recall@50 & Configuration \\
\toprule
WRMF+SGD & 0.2885 & Latent dimensionality: 128 \\
WRMF+ALS & 0.2898 & Latent dimensionality: 128 \\
ItemAE & 0.2862 & Network architecture: $(|\mathcal{U}| \rightarrow256 \rightarrow 128 \rightarrow 256 \rightarrow |\mathcal{U}|)$ \\
\midrule
NCF & 0.2878 & Latent dimensionality: 256; 1 layer NN with size 128 \\
Mult-VAE & 0.2905 & Network architecture: $(|\mathcal{I}| \rightarrow512 \rightarrow 256 \rightarrow 512 \rightarrow |\mathcal{I}|)$ \\
CML & 0.2872 & Latent dimensionality: 256; margin in hinge loss: 10 \\
ItemCF & 0.2191 & Jaccard similarity; number of nearest neighbors: 50 \\
\bottomrule
\end{tabular}
\label{tb:model_config}
\end{table*}
\endgroup

\section{Empirical Studies}\label{sec:expr}
Conducted on a real-world dataset, experiments in this section are divided into two parts. Firstly, we analyze attack transferability from surrogate models (introduced in~\cref{sec:solution}) to different types of victim recommenders. 
Next, we aim to identify the limitations of this adversarially learned injection attack.
Source code and processed dataset are publicly available online\footnote{\url{https://github.com/graytowne/revisit_adv_rec}}. 

\subsection{Setup}
\noindent{\textbf{Dataset.}}
For the real-world dataset, we use Gowalla\footnote{\url{http://snap.stanford.edu/data/loc-Gowalla.html}} constructed by~\citet{cho2011friendship}, containing implicit feedback through user-venue check-ins. It has been widely adopted in previous works~\cite{tang2018caser,bao2012location} for point-of-interest recommendation. Following~\cite{tang2018caser}, we process the raw data by removing cold-start users and items of having less than 15 feedbacks. The processed dataset contains 13.1k users, 14.0k venues (items) and has a sparsity of 99.71\%. We randomly hold 20\% of the data for the test set and use the remaining data for the training set. As there are more than 1 test items, we use Recall@50, instead of hit ratio, to measure recommendation performance.

\noindent{\textbf{Evaluation protocol.}}
To have a fair study for attack transferability under black-box setting, each attacking method generates a fixed number of fake users (1\% of real users, i.e., $|\mathcal{V}|$=0.01$|\mathcal{U}|$=131) that \emph{has the greatest attack impact on the surrogate model}. We then combine each fake data with normal data and let different victim models trained from scratch with the combined poison data.
For the attack performance on Gowalla, we randomly sample 5 items together as a target item set and measure the HR@50 on the target item set (it is considered as a hit if one of these items appears in the ranked list). With more target items involved, the attack performance will be more significant and stable.\enlargethispage{12pt}

\subsection{Analyses on Attack Transferabilities}
In the subsection, we aim to explore the key factors that influence the attack transfer from one surrogate model to other victim models. 
For the attacking methods, we use the ones described in~\cref{sec:solution}. That is, the compared methods are:
\begin{itemize}
    \item \texttt{RandFilter:} A basic attacking method proposed by~\citet{lam2004shilling}. Though the original version is for explicit ratings, we adapt this method on implicit feedback data by having each fake user click the target item(set) and some other randomly chosen items (called filter items). This method serves as a heuristic baseline.
    \item \texttt{WRMF+ALS:} Using WRMF optimized with ALS as surrogate model to craft fake users. Same as~\cite{li2016data, fang2020influence}, only the partial derivative is used as adversarial gradient.
    \item \texttt{WRMF+SGD:} Proposed in~\cref{sec:solution_exact}, this method uses WRMF optimized with Adam as surrogate model. When accumulating adversarial gradients, we approximate the exact adversarial gradient by unrolling 10\% of total training steps.
    \item \texttt{ItemAE:} Proposed in~\cref{sec:solution_approx2}, this method uses item-based autoencoder optimized with Adam as surrogate model. The special design of ItemAE allows us to obtain non-zero partial derivatives. Thus, when accumulating adversarial gradients, we unroll either 0 steps (using only partial derivative) or 10\% of total training steps.
\end{itemize}
For the victim models, we carefully choose the following commonly used recommenders:
\begin{itemize}
    \item \textit{NCF}~\cite{he2017neural} Neural collaborative filtering (NCF) is a popular framework that explore non-linearities in modeling complex user-item interactions. We adopt NeuMF as the instantiation of NCF.
    \item \textit{Mult-VAE}~\cite{liang2018variational} Variational autoencoder with a multinomial likelihood (Mult-VAE) is the state-of-the-art model for recommendation with implicit feedback. It exploits VAE to learn robust user representations and shows much better performances than other factorization-based methods.
    \item \textit{CML}~\cite{hsieh2017collaborative} Collaborative metric learning (CML) minimize the euclidean distance in latent space for a relevant user-item pair and increase the euclidean distance for an irrelevant pair. It is adopted here to see whether \emph{difference in score functions} (i.e., euclidean distance in CML versus dot-product in WRMF and ItemAE) can influence attack transfer.
    \item \textit{ItemCF}~\cite{sarwar2001item} Presumably, all above victim models are based on user/item embeddings, which is a similar properties shared by our surrogate models. Therefore, we choose the item-based collaborative filtering (ItemCF), a classic neighborhood-based approach, to see if attack can transfer to a victim model with \emph{different model type}.
\end{itemize}
To rise reproducibility, in~\cref{tb:model_config} we report the recommendation performance (without attack) and the configuration of each model used in this section. Note that we did not tune each model exhaustively but roughly grid search for the hyperparameters untill a reasonable recommendation performance is reached, because comparing recommendation performance is not our main focus. Besides the aforementioned victim models, we also measure the attack performance when the victim models are identical to the surrogate models (i.e., WRMF and ItemAE).\enlargethispage{12pt}

\begin{figure*}[t!]  %
\centering
    \centering
    \begin{subfigure}[b]{0.45\textwidth}
        \includegraphics[width=\textwidth]{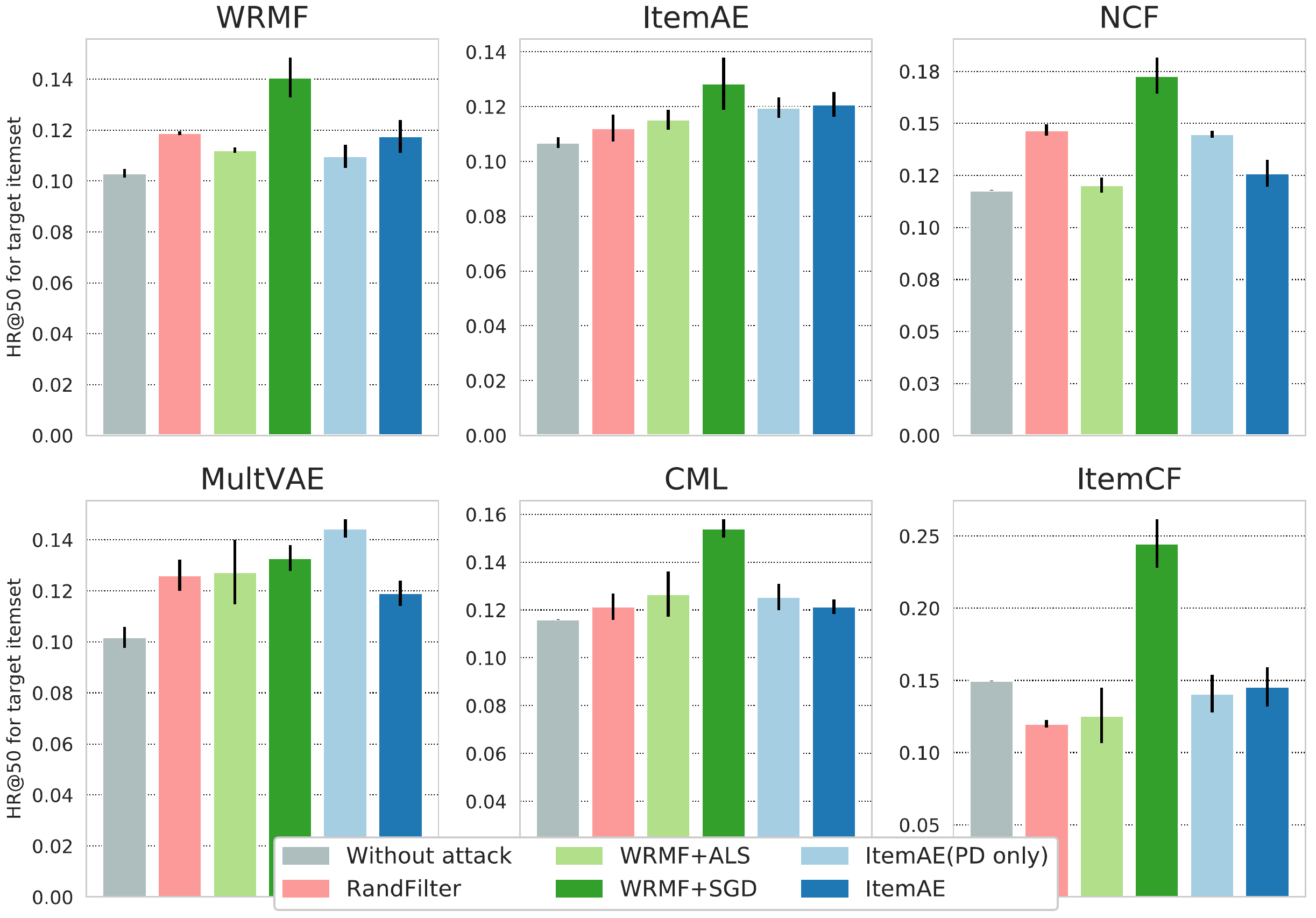}
        \vspace{-1em}
        \caption{Attack performance for different methods.}
        \label{fig:victim_gowalla}
    \end{subfigure}
    ~
    \begin{subfigure}[b]{0.25\textwidth}
        \includegraphics[width=\textwidth]{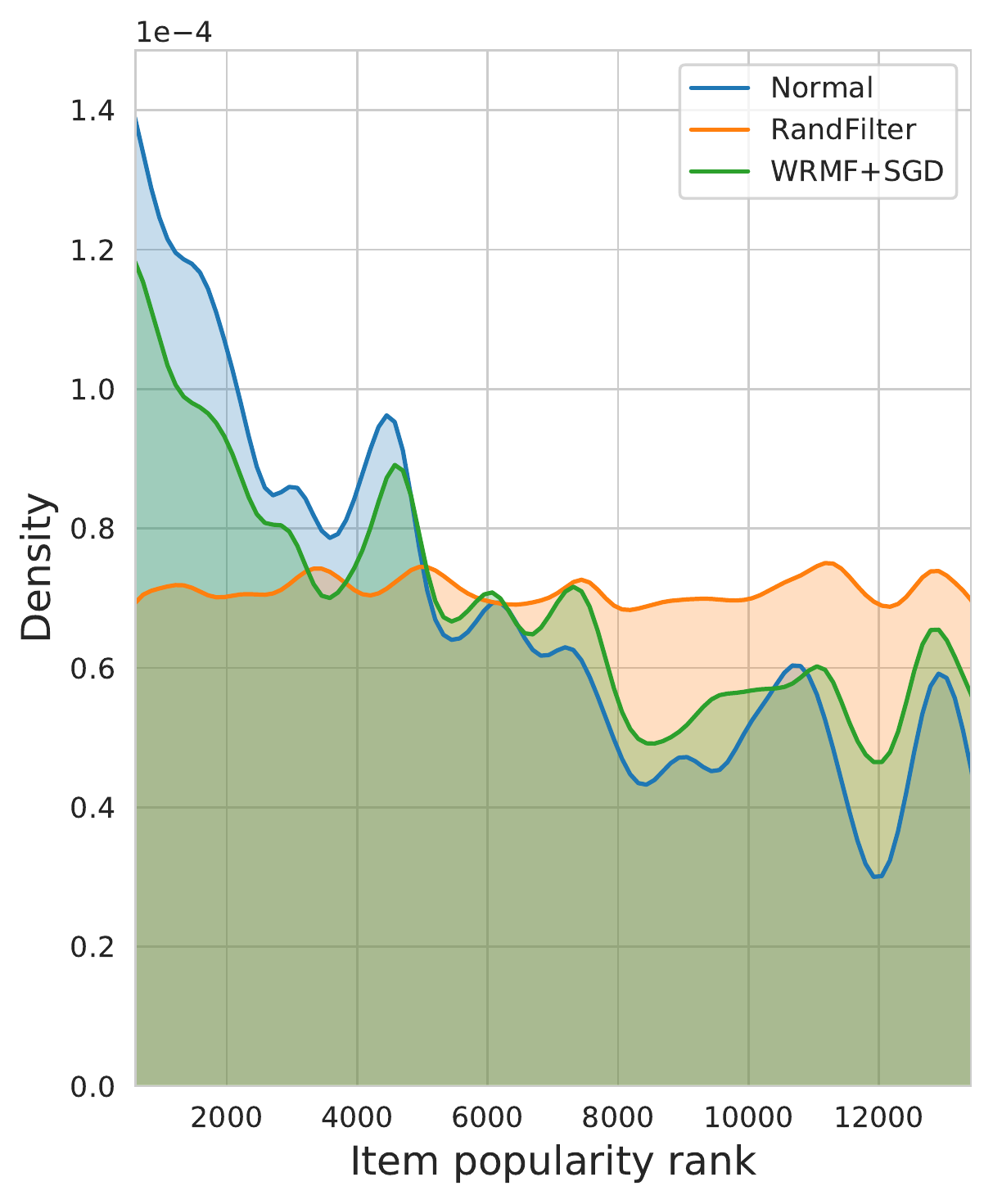}
        \vspace{-1em}
        \caption{Fake data distribution.}
        \label{fig:fake_data_distribution}
    \end{subfigure}
    ~
    \begin{subfigure}[b]{0.3\textwidth}
        \includegraphics[width=\textwidth]{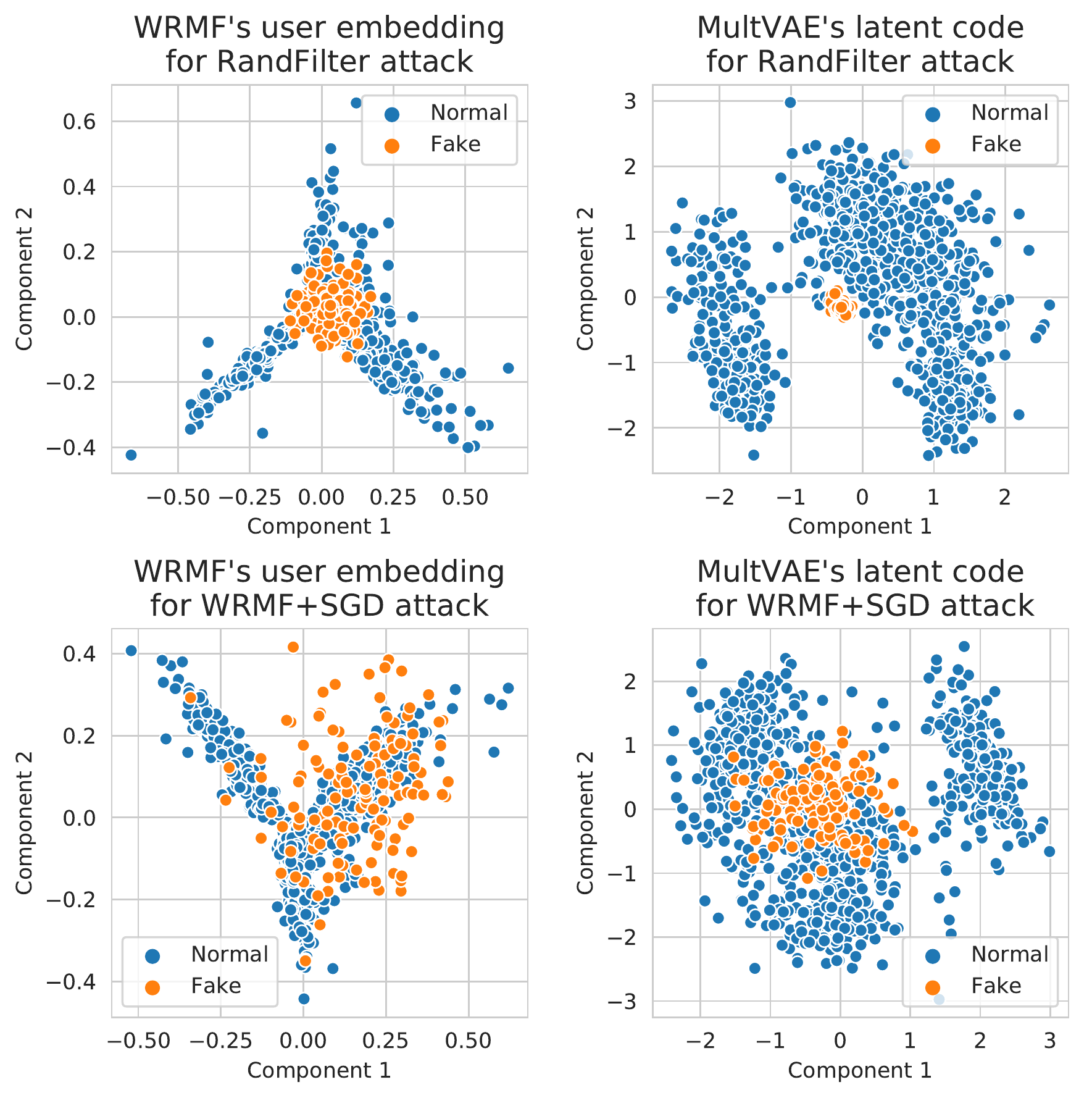}
        \vspace{-1em}
        \caption{Fake users in latent space.}
        \label{fig:fake_data_latent}
    \end{subfigure}
\vspace{-2em}
\caption{(a): On Gowalla dataset, the black-box attack performance for different attacking methods on different victim models. For each result, we report the mean and standard deviation over 4 individual runs with different initializations. (b): In terms of distribution, there's isn't large discrepancy from the learned fake data and the normal data. (c): Fake users in latent space, PCA is used to project user embeddings to a 2-dimensional latent space.}
\label{fig:victim_gowalla}
\end{figure*}

\begingroup
\renewcommand{\arraystretch}{0.9}
\setlength{\tabcolsep}{8pt}
\begin{table*}[t!]
\caption{Attack performance of \texttt{WRMF+SGD} on each victim model for target item set with different popularity.}
\vspace{-1em}
\label{tb:limitation_cold_item}
\begin{tabular}{r|c|ccccccc}
\toprule
\textbf{Target item} & \multirow{2}{*}{\textbf{Method}} & \multicolumn{6}{c}{\textbf{HR@50 for target item set}}\\
\textbf{popularity} & & WRMF & ItemAE & NCF & Mult-VAE & CML & ItemCF\\
\toprule

\multirow{2}{*}{head} & \texttt{Clean} 
& 0.1014 & 0.1043 & 0.1177 & 0.0957 & 0.1159 & 0.1498\\
& \texttt{WRMF+SGD} 
& 0.1405 & 0.1268 & 0.1254 & 0.1682 & 0.1595 & 0.2554\\
\midrule

\multirow{2}{*}{upper torso} & \texttt{Clean} 
& 0.0345 & 0.0371 & 0.0287 & 0.0225 & 0.0251 & 0.0218\\
& \texttt{WRMF+SGD} 
& 0.0590 & 0.0457 & 0.0577 & 0.0371 & 0.0482 & 0.0807\\
\midrule

\multirow{2}{*}{lower torso} & \texttt{Clean} 
& 0.0070 & 0.0064 & 0.0106 & 0.0093 & 0.0101 & 0.0167\\
& \texttt{WRMF+SGD} 
& 0.0287 & 0.0211 & 0.0264 & 0.0123 & 0.0252 & 0.0227\\
\midrule

\multirow{2}{*}{tail} & \texttt{Clean} 
& 0.0005 & 0.0005 & 0.0018 & 0.0025 & 0.0019 & 0.0069\\
& \texttt{WRMF+SGD} 
& 0.0183 & 0.0139 & 0.0175 & 0.0114 & 0.0174 & 0.0194\\
\bottomrule
\end{tabular}
\end{table*}
\endgroup
\cref{fig:victim_gowalla} shows the results for attack performance of each method under black-box setting. The sampled items are constrained to be popular items, thus has a relatively high HR@50 before attack happens. 
As expected, when \textit{WRMF} and \textit{ItemAE} are selected as the victim models, they are affected most when the same models (i.e., \texttt{WRMF+SGD} and \texttt{ItemAE}) are used as the surrogate models. 
For the heuristic method \texttt{RandFilter}, it could achieve the malicious goal sometimes but gives unstable attack performances across different victim models.
The attack generated by \texttt{WRMF+SGD} transfers well to all other models, but the results are much worse in most cases when using \texttt{WRMF+ALS}, which is adopted in existing works.
As for ItemAE, unrolling more steps (\texttt{ItemAE}) does not give better attack performance and not provide better attack transferability in most cases than only using the partial derivative (\texttt{ItemAE(PD)}). Also, \texttt{ItemAE(PD)} shows significant attack influence when the victim model is \textit{Mult-VAE}, the common structure of the two models may be reason.
Lastly, we found that the difference in score functions (i.e., euclidean distance in \textit{CML} versus dot-product in our surrogate models) does not affect the attack too much. This finding suggests on the latent space, the injected fake users actually `pull' the target items towards normal users, such that both cosine distance (from dot-product) and euclidean distance become smaller.
However, the difference in the choice of the victim model (i.e., using \textit{ItemCF} as victim model) can deteriorate the attack impact a lot, except for the case when \texttt{WRMF+SGD} is used to generate the attack. 

\subsection{Limitations of the Attack}
As mentioned, only knowing the severity of the attack is not always useful for defending against the attack. In this subsection, we discuss two major limitations we identified from the proposed injection attack, with the goal of providing insights to further understand this type of attack and inspire defensive techniques.

\noindent{\textbf{Less effective on cold items.}}
Note that in the previous subsection, we showed the attack effectiveness only on a set of randomly sampled popular item. 
In~\cref{tb:limitation_cold_item}, we present the results for \texttt{WRMF+SGD} attack on various victim models, but this time target item sets are sampled to have different popularities. We define head item as the items with total clicks (\#clicks) above 95 percentile. 
Similar definitions also apply for upper torso (75 percentile < \#clicks < 95 percentile), lower torso (75 percentile < \#clicks < 50 percentile) and tail (\#clicks < 50 percentile). 
From~\cref{tb:limitation_cold_item}, we can see the attack from \texttt{WRMF+SGD}, though still boost the target item sets, is less effective for the target items with less popularity.
In other words, the cold items are much harder to get promoted. Perhaps this is because cold items are farther away from normal users on the latent space, thus brings more difficulties for the attack.

\noindent{\textbf{Learned fake users are detectable.}}
Next, we aim to find if there are any clues of the learned fake users.
When target items are head items, we first take a look on what items are clicked by those learned fake users in terms of the clicked item's popularity and the corresponding density. 
\cref{fig:fake_data_distribution} gives the results for \texttt{RandFilter} and \texttt{WRMF+SGD} attack. For reference, we also plot for a sub-sample of 500 normal users. 
From the figure, we can see the clicked items from \texttt{RandFilter} attack have totally random popularity, as expected. But the fake user distribution of \texttt{WRMF+SGD} attack has marginal difference from normal users (labeled as Normal), suggesting the difficulty of identifying the fake data from the distribution discrepancy.
We then alter to seek clues from latent space with the help of PCA. In~\cref{fig:fake_data_latent}, we plot the fake users from \texttt{RandFilter} and \texttt{WRMF+SGD} attack in the latent space of \textit{WRMF} and \textit{MultVAE}. From the first row of~\cref{fig:fake_data_latent}, we verified the claim in~\cite{burke2015robust} that fake users generated with heuristic approach (here the \texttt{RandFilter}) can self-forming clusters in latent space. Notably the fake users from \texttt{WRMF+SGD} attack, although not form cluster in the latent space of \textit{WRMF}, is suspect in the latent space of \textit{MultVAE}. 
This suggests the learned fake users may still self-form clusters in certain latent spaces, once we are able to identify some fake users, their neighbors in latent space are also suspectable.

\section{Conclusion}
\label{sec:conclu}
In this paper, we revisit the problem of adversarially-learned injection attack against recommender systems. By modeling the attacking problem as a bi-level optimization problem, we point out the issue in existing solutions, propose the exact solution to solve the problem and verify its effectiveness on a synthetic dataset.
Since the exact solution is neither time-efficient nor resource-efficient, we also introduce two approaches to approximate the exact solution.
On a real-world dataset, we conduct extensive experiments to evaluate the transferability and limitations of the attack, with the hope that these empirical analyses can inspire effective defensive strategies in the future.

\begin{acks}
The work of the third author is partially supported by a Discovery Grant from Natural Sciences and Engineering Research Council of Canada.
\end{acks}

\bibliographystyle{ACM-Reference-Format}
\bibliography{ref_kdd19}

\end{document}